\documentclass[11pt]{article}

\usepackage{acl}

\usepackage{times}
\usepackage{latexsym}

\usepackage[T1]{fontenc}

\usepackage[utf8]{inputenc}

\usepackage{microtype}

\usepackage{inconsolata}

\usepackage{graphicx}

\usepackage{url}            % simple URL typesetting
\usepackage{booktabs}       % professional-quality tables
\usepackage{amsfonts}       % blackboard math symbols
\usepackage{nicefrac}       % compact symbols for 1/2, etc.
\usepackage{colortbl}       % \cellcolor for table cells
\usepackage{multirow}       % multi-row table cells
\usepackage{array}          % table column tweaks
\usepackage{tabularx}       % column-width tables for the two-column layout
\usepackage{enumitem}       % list controls
\usepackage{amsmath}        % math
\usepackage{amssymb}        % math symbols
\usepackage{xspace}         % \sys spacing before punctuation
\usepackage{listings}       % verbatim code/prompt blocks with line wrapping
\usepackage[breakable,skins,listings]{tcolorbox}  % styled boxes with listings inside
\lstdefinestyle{prompt}{
  basicstyle=\scriptsize\ttfamily,
  breaklines=true,
  breakatwhitespace=true,
  columns=flexible,
  frame=single,
  framesep=4pt,
  xleftmargin=4pt,
  xrightmargin=4pt,
  upquote=true,
  literate={—}{{---}}1 {–}{{--}}1 {…}{{...}}1 {★}{{*}}1
           {↔}{{<->}}1 {→}{{->}}1 {≠}{{!=}}1 {≥}{{>=}}1 {≤}{{<=}}1
           {“}{{"}}1 {”}{{"}}1 {‘}{{'}}1 {’}{{'}}1
}

\newcommand{\sys}{\textsc{CiteTracer}\xspace}
\newcommand{\added}[1]{#1}

\ifdefined\pdfsuppressptexinfo
\fi

\title{Source or It Didn't Happen: A Multi-Agent Framework for Citation Hallucination Detection}

\author{
  \textbf{Mingzhe Li\textsuperscript{1}},
  \textbf{Zhiqiang Lin\textsuperscript{2}},
  \textbf{Shiqing Ma\textsuperscript{1}}
\\
  \textsuperscript{1}University of Massachusetts Amherst,
  \textsuperscript{2}The Ohio State University
}

\begin{document}
\maketitle
\setlength{\itemsep}{0.5pt}
\setlength{\parsep}{0.5pt}
\setlength{\topsep}{0.5pt}
\setlength{\textfloatsep}{1pt} 
\setlength{\floatsep}{0.5pt} 
\setlength{\abovedisplayskip}{0.5pt}
\setlength{\belowdisplayskip}{0.5pt}
\setlength{\abovecaptionskip}{0.5pt}
\setlength{\belowcaptionskip}{0.5pt}
\setlength{\dbltextfloatsep}{6pt}
\begin{abstract}
Large language models are increasingly used in scientific writing, yet they can fabricate citation-shaped references that appear plausible but fail bibliographic verification. Existing detectors often reduce verification to binary found/not-found decisions and rely on brittle parsing or incomplete retrieval, offering little field-level signal to auditors. We reframe citation hallucination detection as taxonomy-aligned field-level adjudication and introduce a $12$-code taxonomy spanning \textsc{Real}, \textsc{Potential}, and \textsc{Hallucinated} citations. Based on this taxonomy, we build \sys, a cascading multi-agent detector that extracts structured citations from PDF and BibTeX, retrieves evidence through cache lookup, URL fetch, scholar connectors, and web search, applies deterministic field matching, and routes ambiguous cases to class-specialist judgers. We release a benchmark of $2{,}450$ synthetic citations built from real seeds with controlled LLM mutations, paired with $957$ real-world fabricated citations drawn from desk-rejected submissions at ICLR 2026 and an anonymous conference. \sys reaches $97.1\%$ accuracy on the synthetic benchmark, with class-level $F_1$ of $97.0$, $95.8$, and $98.5$ for \textsc{Real}, \textsc{Potential}, and \textsc{Hallucinated}, respectively, and detects $97.1\%$ of fabrications on the real-world set without abstaining. Code: \url{https://github.com/aaFrostnova/CiteTracer}.
\end{abstract}

\section{Introduction}
\label{sec:intro}

\begin{figure*}[t]
  \centering
  \includegraphics[width=1.0\textwidth]{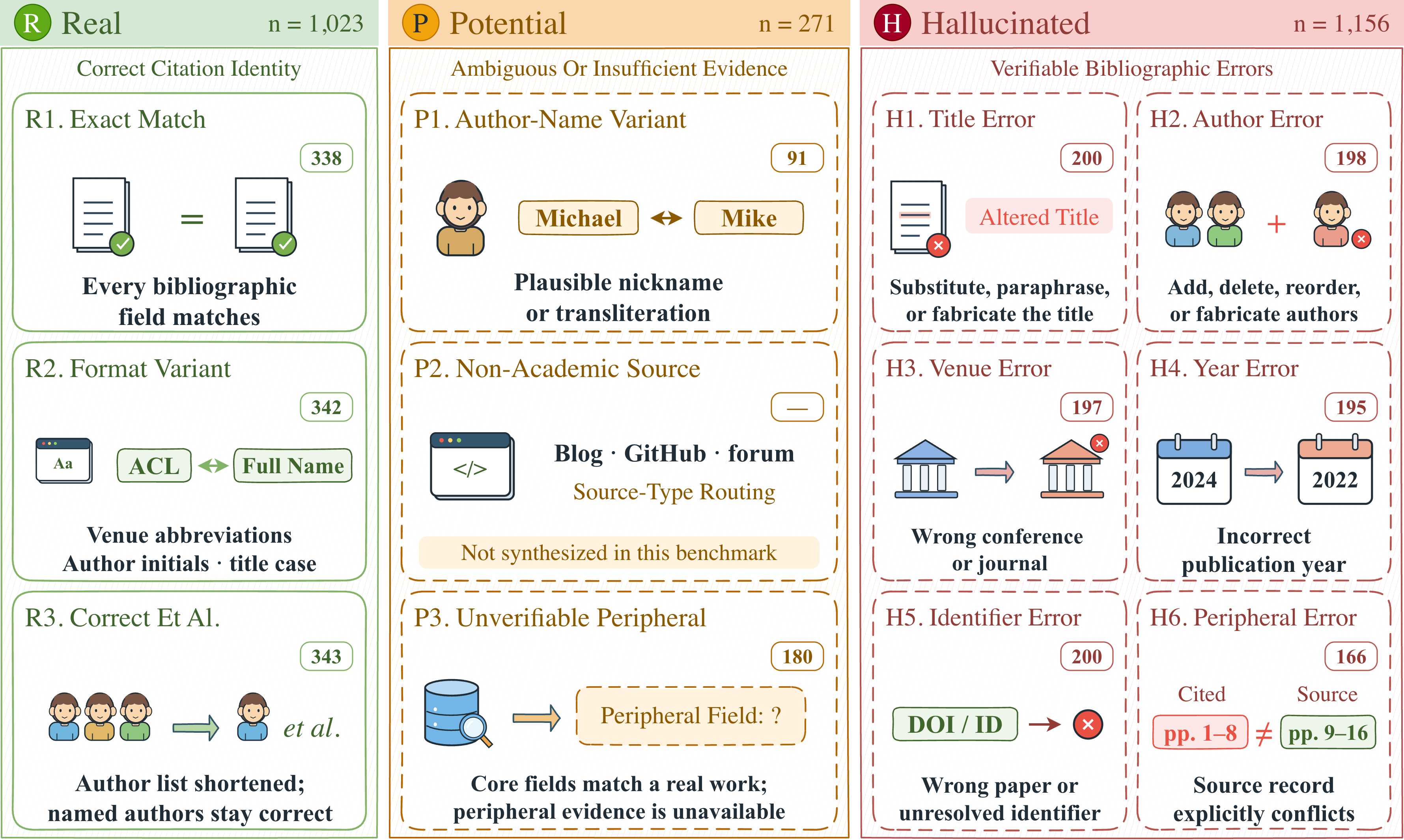}
  \caption{The 12-code citation hallucination taxonomy.}
  \label{tab:taxonomy}
\end{figure*}

Citations are the infrastructure of scientific communication: they justify claims, allocate scholarly credit, and trace the chain of evidence behind every paper~\cite{waltman2016review}.
Within this broader notion of citation integrity, bibliographic integrity asks whether a cited entry's title, authors, venue, year, and identifiers actually correspond to a real publication~\citep{yuan2026citeaudit}.
A bibliographic-level error denies the original authors their credit, breaks reproducibility because the metadata no longer leads back to a retrievable source, and propagates downstream as search engines surface the fabricated entry~\citep{rekdal2014academic,sarol2024assessing}.

Large language models are now deeply embedded in the research workflow, especially in academic writing, where they help generate ideas, polish exposition, and draft submission text. This shift introduces a new bibliographic failure mode: an LLM can rely on distributional patterns in text to produce citation-shaped entries with hallucinated or mismatched fields, such as an incorrect title, a nonexistent author, or a venue that does not correspond to the cited work~\citep{yuan2026citeaudit}. This risk follows from the broader problem of hallucination, but citations make the failure especially consequential: they are high-stakes factual claims whose fields should be externally verifiable, yet LLMs are highly fluent at producing references that appear plausible by construction~\citep{walters2023fabrication,chelli2024hallucination}.
Hallucinated citations range from incorrect metadata on real papers, to entries that mix real and fabricated fields, to entirely nonexistent publications, and they call for different auditor responses (correction, rejection, or uncertainty) rather than a single binary judgment. The problem is now operational at the venue level: ICLR 2026 chairs assembled a desk-reject queue of more than $600$ submissions flagged for fabricated references, and ICML and ACM CCS have announced similar policies for the 2026 cycle~\citep{sakoi2026hallucitation,gptzero_iclr2026,register2026neurips_hallucinations}.

Existing detectors miss this failure surface in two specific ways.
First, they collapse verification into a binary verdict with no class for unresolved evidence.
Commercial and academic auditors~\citep{citely,swanref,citecheck,refcheck_ai,yuan2026citeaudit} emit only a binary Real-or-Fake verdict~\citep{janse2025ai}, collapsing the ambiguous middle ground into a yes/no signal, while open tools key the verdict on title and author~\citep{hallucinator} or audit only five fields~\citep{gptzero} (Section~\ref{sec:related}).
Second, PDF input compounds the gap: their reference parsers drop entries, mis-segment author and title spans, and occasionally hallucinate fields of their own, so the verifier inherits a corrupted input before any auditing happens.

To address these limitations, we introduce a comprehensive
citation hallucination benchmark and \sys, a multi-agent
verification framework.
\label{sec:intro:taxonomy}
Our benchmark defines a 12-code taxonomy spanning
\textsc{Real}, \textsc{Potential}, and \textsc{Hallucinated}
citations, with controlled field-level mutations of
authentic references.
\sys combines layout-aware PDF extraction, cascading
multi-source retrieval, deterministic field matching,
and class-specialist judgment to produce interpretable
field-level verdicts supported by external evidence.

\added{On the $2{,}450$-citation synthetic benchmark,
\sys achieves $97.55\%$ three-class accuracy on BibTeX
inputs and $97.10\%$ fine-grained diagnostic accuracy.
It also achieves $97.50\%$ accuracy on BIOMED-1K
without domain-specific modifications and identifies
$97.1\%$ of known fabricated references in real-world
submissions.
Further ablations demonstrate the contributions of
specialized judgment and field-level matching.}

Our contributions are summarized as follows:
\begin{itemize}[leftmargin=1.5em,itemsep=2pt,topsep=2pt]
  \item We introduce a 12-code citation taxonomy and
  a $2{,}450$-citation benchmark with controlled
  field-level mutations.

  \item We propose \sys, a multi-agent citation auditor
  combining layout-aware extraction, cascading retrieval,
  field matching, and class-specialist judgment.

  \item We evaluate \sys against five baselines on
  PDF and BibTeX inputs, demonstrating strong
  detection performance, cross-domain generalization,
  and the contributions of individual components.
\end{itemize}

\section{Related Work}
\label{sec:related}

\noindent \textbf{Hallucination in Academic Writing.}
Large language models hallucinate factual content even when surface fluency is maintained, a failure mode characterized across model families, training regimes, and deployment settings in recent surveys~\citep{huang2025survey,tonmoy2024comprehensive,rahman2026hallucination} and in zero-resource detection work such as SelfCheckGPT~\citep{selfcheckgpt}.
Fabricated references are a prominent instance~\citep{walters2023fabrication,chelli2024hallucination,sakoi2026hallucitation}: chairs at NeurIPS 2025~\citep{gptzero_neurips,register2026neurips_hallucinations}, ICLR 2026~\citep{gptzero_iclr2026}, and ACM CCS 2026~\citep{ccs2026_transparency} have documented them at venue scale.

\noindent \textbf{Citation Hallucination Detection.}
Commercial auditors such as Citely~\citep{citely}, SwanRef~\citep{swanref}, CiteCheck~\citep{citecheck}, and RefCheck-AI~\citep{refcheck_ai} return only a binary Real-or-Fake verdict~\citep{janse2025ai} without naming the wrong field. CiteAudit~\citep{yuan2026citeaudit} combines a taxonomy of bibliographic errors with binary citation verification.
Hallucinator~\citep{hallucinator} consults more than ten bibliographic databases but keys the verdict on title and author, and GPTZero's hallucination mode~\citep{gptzero} audits only five fields (title, author, date, URL, publisher), accepts only PDF input, and requires a paid subscription.
None of these systems separates the \textsc{Potential} middle ground (nickname variants, non-academic sources, peripheral metadata gaps) from fabrications. In contrast, \sys represents unresolved evidence explicitly through the \textsc{Potential} class and is evaluated on fine-grained diagnostic codes.

\section{Benchmark}
\label{sec:data}

Existing citation auditors are largely closed-source and report opaque metrics, and the CiteAudit benchmark~\citep{yuan2026citeaudit} evaluates binary verdicts, so no open benchmark yet compares methods on three-class, field-level ground truth.
We close this gap with a $2{,}450$-citation synthetic benchmark grounded in real bibliographies and a $957$-citation real-world test set drawn from the ICLR 2026 desk-reject queue ($807$ citations) and another anonymous conference ($150$ citations). Full construction and per-code details are deferred to Appendix~\ref{sec:appendix:benchmark}.

\noindent\textbf{Taxonomy.}
A bibliographic citation decomposes into a fixed set of fields (title, authors, venue, year, identifiers, peripheral metadata), and the appropriate auditor response depends on which field is wrong and whether the error can be verified externally.
We define $12$ fine-grained codes grouped into three auditor-facing classes (\autoref{tab:taxonomy}).
\textsc{Real} (\textsc{R1}--\textsc{R3}) covers exact matches and normalizable formatting variants such as venue abbreviations, author initials, and \emph{et al.} truncation.
\textsc{Hallucinated} (\textsc{H1}--\textsc{H6}) localizes a single bibliographic error to one field: title (\textsc{H1}), authors (\textsc{H2}), venue (\textsc{H3}), year (\textsc{H4}), identifier (\textsc{H5}), or peripheral metadata (\textsc{H6}).
\textsc{Potential} (\textsc{P1}--\textsc{P3}) buffers auditor-ambiguous cases: nickname or transliteration variants (\textsc{P1}), non-academic sources whose existence cannot be verified through bibliographic indices (\textsc{P2}), and peripheral fields that no public source records for the cited paper (\textsc{P3}).
Per-field localization gives the benchmark its diagnostic value: a wrong title and a wrong DOI on otherwise identical seeds correspond to two distinct error modes that require different auditor corrections.

\noindent\textbf{Construction.}
We draw seed BibTeX entries from open-access bibliographic repositories (e.g., DBLP, arXiv, ACL) across $50$ recent ML and CS papers, prioritizing entries that populate the largest set of fields.
For every non-\textsc{R1} code we apply a code-specific mutation operator that touches a documented set of fields and leaves the rest of the seed identical: an LLM-driven generator proposes a candidate value, and a deterministic post-processor enforces the operator's field schema.
We do not include synthetic \textsc{P2} cases because \textsc{P2} is defined by source type rather than bibliographic-field correctness: any clearly non-academic citation, such as a blog post, GitHub repository, or forum thread, is directly routed to \textsc{P2}, making it a routing case rather than a challenging verification case.
Each synthetic entry passes three independent checks before it enters the benchmark: a round-trip audit on operator diffs, a verifiability check on every \textsc{R1} and \textsc{P3} entry, and an author-curated boundary review on every \textsc{P1} substitution.
These checks retain $2{,}450$ taxonomy-labeled instances out of $3{,}100$ generated entries, with per-code counts reported alongside each code in \autoref{tab:taxonomy}.
\added{For cross-domain evaluation, we additionally constructed BIOMED-1K: $1{,}000$ biomedical citations seeded from PubMed, Europe PMC, and bioRxiv, using the same taxonomy and mutation operators without changing the verification pipeline or prompts.}

\noindent\textbf{Real-world test set.}
We additionally collect two real-world slices on which fabrications were flagged by the venue's own chairs.
The first slice contains $807$ citations from $647$ \textit{ICLR 2026} submissions that the program chairs desk-rejected for fabricated references%
\footnote{\url{https://openreview.net/group?id=ICLR.cc/2026/Conference\#tab-desk-rejected-submissions}}.
The second slice contains $150$ citations from $41$ desk-rejected submissions at an anonymous conference.
Every entry in both slices carries the chairs' verdict and the cited bibliographic record, so synthetic-set numbers can be cross-checked against fabrications two different venues actually rejected.

\section{Methodology}
\label{sec:method}
\begin{figure*}[t]
  \centering
  \includegraphics[width=1.0\textwidth]{figs/overview.png}
  \caption{Overview of \sys. (1) The \emph{Reference Extractor} parses each citation block into a structured field-level record. (2) The \emph{Cascading Evidence Collector} walks a memory cache, URL fetch, eight Scholar Connectors, and web search. (3) The \emph{Field Matcher} compares the record against the evidence field by field. (4) \emph{Class-specialist Judgers} adjudicate ambiguous cases and emit a taxonomy-aligned verdict with the offending fields and reasons.}
  \label{fig:overview}
\end{figure*}

In this section, we introduce \sys, an end-to-end agentic framework that turns citation hallucination detection into per-citation, per-field verdicts an auditor can act on.
Formally, \sys maps an input paper $P$ to $\sys(P)=\{(r_i, y_i, \Delta_i, \mathcal{S}_i)\}_{i=1}^{N}$, where $r_i$ is the $i$-th structured citation record, $y_i$ its taxonomy-aligned verdict, $\Delta_i$ the set of offending field spans, and $\mathcal{S}_i$ the bibliographic sources supporting the decision. \autoref{fig:overview} provides an overview of \sys{}.

\subsection{Reference Extractor}
\label{sec:method:parser}

The Reference Extractor takes a paper as input and produces a list of canonical citation records, with every bibliographic field a downstream verifier might check.
This step is challenging because citation extraction still requires character-level precision under realistic PDF layouts.
Although modern OCR systems can detect bibliography regions and citation blocks, their transcriptions may still contain subtle character-level errors, especially for author names, venue abbreviations, page numbers, and identifiers.
Moreover, bibliography styles vary widely across papers, and even references within the same paper may exhibit different surface formats.
As a result, purely rule-based extraction is often brittle and difficult to scale across bibliography styles, and learning-based approaches such as soft-constrained citation field extractors trained on the UMass Citations corpus~\citep{AnzarootPBM14,UMassCitations} still leave residual character-level errors that propagate into downstream verification.

To address these issues, we use the OCR model as a high-recall citation-block proposer rather than as the final parser.
Let $\mathcal{M}_{\mathrm{ocr}}$ denote the OCR model.
Given the bibliography region $P_{\mathrm{bib}}$ of an input paper $P$, the OCR model returns citation blocks together with their initial transcriptions:
\[
\{(B_k,T_k)\}_{k=1}^{K}
=
\mathcal{M}_{\mathrm{ocr}}(P_{\mathrm{bib}}),
\]
where $B_k$ is the page-level region of the $k$-th detected citation block, and $T_k$ is its OCR transcription.
We then introduce a parsing agent as a second safeguard.
Let $\mathcal{A}_{\mathrm{Parser}}$ denote the parsing agent.
For each detected citation block, the agent takes the cropped block image and its OCR transcription as input, rechecks the extracted text against the visual evidence, and directly extracts structured bibliographic fields.
Formally, let $\mathcal{F}$ denote the set of bibliographic fields to be verified, including title, authors, venue, year, DOI, pages, publisher, location, and URL.
For the $k$-th detected citation block, the parsing agent produces a provisional structured citation record:
\[
r_k
=
\mathcal{A}_{\mathrm{Parser}}(P[B_k],T_k)
=
\{(f,v_{k,f}) \mid f \in \mathcal{F}\},
\]
where $v_{k,f}$ is the extracted value of field $f$ from the $k$-th detected citation block.
This crop-level rechecking allows the extractor to repair OCR errors without relying on rigid hand-crafted rules for specific bibliography styles. Some references may be split across a column boundary or a page boundary, so a detected citation block does not always correspond to a complete reference.
In these boundary cases, the parsing agent identifies continuation blocks and merges their visual-textual evidence before finalizing the structured record.
This boundary repair step allows the extractor to recover references that are fragmented across columns or pages.
The final output of the Reference Extractor is the set of structured citation records
\(
\mathcal{R}(P)=\{r_i\}_{i=1}^{N}
\),
where $N$ is the number of finalized references after boundary repair.

\subsection{Cascading Evidence Collector}
\label{sec:method:retrieval}

The Cascading Evidence Collector takes a structured citation record $r_i$ and returns a ranked list of candidate matches together with the bibliographic evidence supporting each match.
This step is challenging because citation verification must balance retrieval cost against source coverage.
Many citations can be resolved by cheap signals, such as previously verified records or explicit DOI/arXiv links, but long-tail references may only appear in specialized bibliographic sources or unstructured web pages.
As a result, querying every source for every citation wastes connector calls on the easy majority, while relying on a single source leaves biomedical papers, ACL Anthology entries, workshop papers, and non-standard web references uncovered.

To address this trade-off, we use a four-stage retrieval cascade ordered from cheapest to most general: \textbf{Memory}, \textbf{URL Fetch}, \textbf{Scholar Connectors}, and \textbf{Web Search}.
The first stage, \textbf{Memory}, queries a cache initialized from an offline DBLP mirror and updated with every newly verified \textsc{Real} citation, in the spirit of long-term memory layers proposed for production agent systems~\citep{mem0}.
It returns previously seen candidate records at near-zero cost.
The second stage, \textbf{URL Fetch}, is triggered when the citation contains explicit links such as a DOI, arXiv URL, or publisher landing page.
The Web Agent follows each URL and extracts structured metadata, so this stage produces evidence from direct citation links rather than from a general query.

The third stage, \textbf{Scholar Connectors}, sends the Scholar Agent to query multiple public bibliographic sources in parallel.
This parallel fan-out keeps latency bounded while covering both general computer science literature and domain-specific sources.
The final stage, \textbf{Web Search}, uses the Web Agent again, but now with a search query generated from the citation record rather than a direct URL, in the spirit of multi-agent systems that collect evidence from open-web sources for misinformation detection and structured data acquisition~\citep{tian2024web,ma2025autodata}.
It retrieves raw web summaries or pages and extracts candidate bibliographic records when structured sources miss.

The cascade stops on a \emph{verdict}.
After each stage, the Field Matcher and Class-Specialist Judgers (Sections~\ref{sec:method:matcher} and~\ref{sec:method:judges}) examine the cumulative \emph{evidence bundle} $\mathcal{E}_i$, the union of candidate records collected by every stage tried so far, and emit a citation-level verdict in \{\textsc{Real}, \textsc{Potential}, \textsc{Hallucinated}\}.
The cascade stops at the first stage whose evidence supports a \textsc{Real} verdict and returns that verdict immediately, skipping the remaining stages.

\subsection{Field Matcher}
\label{sec:method:matcher}

The Field Matcher takes a structured citation record $r_i$ and its evidence bundle $\mathcal{E}_i$ as input, and emits a field-level status profile for downstream judgers.
This step is necessary because citation correctness is often field-dependent: a citation may match the retrieved evidence on title and year, but disagree on authors, venue, DOI, or peripheral metadata.
A citation-level similarity score would hide these differences, whereas field-level matching exposes which parts of the reference are supported by evidence.
The challenge is to avoid unnecessary LLM calls on the easy majority while still handling residual cases that require flexible reasoning.
To address this, the Field Matcher uses two stages.
The first stage is a deterministic rule matcher, which applies field-specific normalizers and supports early exit.
The second stage is a Matcher Agent, which is invoked only when deterministic rules cannot fully resolve the citation.

For the deterministic stage, let $\nu_f(\cdot)$ denote the rule-based normalizer for field $f$.
These normalizers only encode high-confidence, reproducible transformations, such as case folding, punctuation removal, DOI canonicalization, page-range normalization, author-name normalization that preserves author-list order, and known venue abbreviations.
Given the extracted value $v_{i,f}$ of citation $r_i$ and the value $u_{e,f}$ of candidate evidence $e \in \mathcal{E}_i$, the rule matcher assigns a deterministic field status $m^{\mathrm{rule}}_{i,e,f}$: \textsc{match} if $\nu_f(v_{i,f})=\nu_f(u_{e,f})$, \textsc{missing} if either value is empty, and \textsc{mismatch} otherwise.
If some candidate $e \in \mathcal{E}_i$ has $m^{\mathrm{rule}}_{i,e,f}=\textsc{match}$ for every field $f$ present in $r_i$, the matcher exits early and treats the citation as a deterministic \textsc{Valid} case. Otherwise, the case is passed to the Matcher Agent.
Let $\mathcal{A}_{\mathrm{Matcher}}$ denote the Matcher Agent.
Unlike the deterministic normalizer, the Matcher Agent does not merely canonicalize strings but examines the citation, the retrieved evidence, and the rule-based status pattern to produce a residual field-status profile:
\[
\mathbf{m}_i
=
\mathcal{A}_{\mathrm{Matcher}}
\left(
r_i,\mathcal{E}_i,\{m^{\mathrm{rule}}_{i,e,f}\}
\right).
\]
The output $\mathbf{m}_i$ records, for each audited field, whether the residual discrepancy is best explained by a normalizable variation, missing candidate metadata, missing reference metadata, or a true field contradiction.
For example, the Matcher Agent may label an author mismatch as an initial-only variant, a venue mismatch as match after abbreviation, or a publisher/page field as candidate missing.
This residual field-status profile is then passed to the Class-Specialist Judgers for taxonomy-level adjudication.

\subsection{Class-Specialist Judgers}
\label{sec:method:judges}

The Class-Specialist Judgers adjudicate cases that cannot be fully resolved by deterministic field matching and emit a final taxonomy-aligned verdict for each citation.
This step is challenging because different error classes require different decision logic.
For example, format variations such as author initials should be treated differently from missing candidate metadata, and both are different from cases where the retrieved evidence contradicts the cited title, year, DOI, or venue.
A single general-purpose judge over all taxonomy codes can easily become miscalibrated because it must apply different evidence thresholds across \textsc{Real}, \textsc{Potential}, and \textsc{Hallucinated} cases.

To address this issue, we use class-specialist judgers instead of one monolithic judge.
The routing decision is based on the field-status profile produced by the Field Matcher.
Let $\mathbf{m}_i$ denote the final field-level status profile for citation $r_i$, and let $\mathcal{E}_i$ denote its retrieved evidence bundle.
A router $\rho$ selects the specialist judger $J_i=\rho(r_i,\mathcal{E}_i,\mathbf{m}_i)$ from the set $\mathcal{J}_{\mathrm{cls}}$ of class-specialist judgers according to the residual field pattern: normalizable residual cases go to the Valid Judger, ambiguous but plausible cases to the Potential Judger, and evidence-contradicting or evidence-absent cases to the Hallucinated Judger.
The selected judger returns the final decision $(y_i,\Delta_i,\mathcal{S}_i)=J_i(r_i,\mathcal{E}_i,\mathbf{m}_i)$, where $y_i$ is the final taxonomy code, $\Delta_i \subseteq \mathcal{F}$ the offending or unresolved fields, and $\mathcal{S}_i$ the supporting evidence.

\section{Evaluation}
\label{sec:exps}
\begin{table*}[t]
  \centering
  \footnotesize
\setlength{\tabcolsep}{3.0pt}
    \renewcommand{\arraystretch}{1.08}
    \setlength{\aboverulesep}{0.30ex}
    \setlength{\belowrulesep}{0.35ex}
  \resizebox{\textwidth}{!}{%
  \begin{tabular}{@{}ll|r|rrr|rrr||r|rrr|rrr@{}}
    \toprule
    \multirow{3}{*}{Method} & \multirow{3}{*}{True} & \multicolumn{7}{c||}{PDF} & \multicolumn{7}{c}{BibTeX} \\
    \cmidrule(lr){3-9}\cmidrule(l){10-16}
     & & \multirow{2}{*}{$N_c$} & \multicolumn{3}{c|}{\added{Predicted label}} & \multicolumn{3}{c||}{Metrics (\%)}
      & \multirow{2}{*}{$N_c$} & \multicolumn{3}{c|}{\added{Predicted label}} & \multicolumn{3}{c}{Metrics (\%)} \\
    \cmidrule(lr){4-6}\cmidrule(lr){7-9}\cmidrule(lr){11-13}\cmidrule(l){14-16}
     & & & \added{R} & \added{P} & \added{H} & Precision & Recall & $F_1$ & & \added{R} & \added{P} & \added{H} & Precision & Recall & $F_1$ \\
    \midrule
    \multirow{3}{*}{GPT-5.5} & R & 1,015 & \cellcolor[HTML]{B9D6C4}\added{957} & \cellcolor[HTML]{B7D4C3}\added{8} & \cellcolor[HTML]{D2E6D6}\added{50} & 85.8 & 94.3 & 89.9 & 1,023 & \cellcolor[HTML]{BAD6C5}\added{958} & \cellcolor[HTML]{B5D3C1}\added{5} & \cellcolor[HTML]{DAE8D8}\added{60} & 94.6 & 93.6 & 94.1 \\
     & P & 245 & \cellcolor[HTML]{F6D0B0}\added{54} & \cellcolor[HTML]{F7D6B8}\added{39} & \cellcolor[HTML]{F5CAA8}\added{152} & 69.6 & 15.9 & 25.9 & 271 & \cellcolor[HTML]{E3EADA}\added{19} & \cellcolor[HTML]{F9E1C7}\added{78} & \cellcolor[HTML]{F5CAA8}\added{174} & 91.8 & 28.8 & 43.8 \\
     & H & 1,132 & \cellcolor[HTML]{F2EBD7}\added{104} & \cellcolor[HTML]{B7D4C3}\added{9} & \cellcolor[HTML]{BFD9C8}\added{1,019} & 83.5 & 90.0 & 86.6 & 1,156 & \cellcolor[HTML]{C6DECD}\added{36} & \cellcolor[HTML]{B3D2C0}\added{2} & \cellcolor[HTML]{B8D5C3}\added{1,118} & 82.7 & 96.7 & 89.2 \\
    \midrule
    \multirow{3}{*}{Claude 4.7 Opus} & R & 1,015 & \cellcolor[HTML]{BBD7C5}\added{948} & \cellcolor[HTML]{D0E5D5}\added{47} & \cellcolor[HTML]{BFD9C8}\added{20} & \added{80.0} & \added{93.4} & \added{86.2} & 1,023 & \cellcolor[HTML]{BED9C7}\added{949} & \cellcolor[HTML]{D2E6D6}\added{50} & \cellcolor[HTML]{C1DBCA}\added{24} & \added{86.4} & \added{92.8} & \added{89.4} \\
     & P & 245 & \cellcolor[HTML]{F6CDAC}\added{69} & \cellcolor[HTML]{F8DABC}\added{49} & \cellcolor[HTML]{F5CAA8}\added{127} & \added{31.6} & \added{20.0} & \added{24.5} & 271 & \cellcolor[HTML]{F8D7BA}\added{44} & \cellcolor[HTML]{FAE2C9}\added{83} & \cellcolor[HTML]{F5CAA8}\added{144} & \added{46.4} & \added{30.6} & \added{36.9} \\
     & H & 1,132 & \cellcolor[HTML]{F9DDC3}\added{168} & \cellcolor[HTML]{D4E6D6}\added{59} & \cellcolor[HTML]{CDE3D3}\added{905} & \added{86.0} & \added{79.9} & \added{82.9} & 1,156 & \cellcolor[HTML]{F2EBD7}\added{106} & \cellcolor[HTML]{CCE2D2}\added{46} & \cellcolor[HTML]{C2DCCB}\added{1,004} & \added{85.7} & \added{86.9} & \added{86.3} \\
    \midrule
    \multirow{3}{*}{Gemini 3.1 Pro} & R & 1,015 & \cellcolor[HTML]{BFD9C8}\added{913} & \cellcolor[HTML]{C7DECE}\added{33} & \cellcolor[HTML]{E1EADA}\added{69} & 61.5 & 90.0 & 73.0 & 1,023 & \cellcolor[HTML]{C1DBC9}\added{905} & \cellcolor[HTML]{F1EBD8}\added{92} & \cellcolor[HTML]{C2DBCA}\added{26} & 57.8 & 88.5 & 69.9 \\
     & P & 245 & \cellcolor[HTML]{F5CAA8}\added{168} & \cellcolor[HTML]{F7D3B4}\added{28} & \cellcolor[HTML]{F7D1B2}\added{49} & 20.7 & 11.4 & 14.7 & 271 & \cellcolor[HTML]{F5CAA8}\added{213} & \cellcolor[HTML]{F7D6B8}\added{43} & \cellcolor[HTML]{D7E7D7}\added{15} & 14.0 & 15.9 & 14.9 \\
     & H & 1,132 & \cellcolor[HTML]{F5CAA8}\added{404} & \cellcolor[HTML]{E0E9D9}\added{74} & \cellcolor[HTML]{EEEEDD}\added{654} & 84.7 & 57.8 & 68.7 & 1,156 & \cellcolor[HTML]{F5CAA8}\added{448} & \cellcolor[HTML]{F9DCC2}\added{173} & \cellcolor[HTML]{FCEEDB}\added{535} & 92.9 & 46.3 & 61.8 \\
    \midrule
    \multirow{2}{*}{Hallucinator$^{\dagger}$} & R & 579 & \cellcolor[HTML]{CBE1D1}\added{480} & \cellcolor[HTML]{F6F5F1}\textemdash & \cellcolor[HTML]{F7D3B5}\added{99} & 44.9 & 82.9 & 58.3 & 1,023 & \cellcolor[HTML]{B8D5C3}\added{992} & \cellcolor[HTML]{F6F5F1}\textemdash & \cellcolor[HTML]{C5DDCD}\added{31} & 53.2 & 97.0 & 68.7 \\
     & H & 800 & \cellcolor[HTML]{F5CAA8}\added{588} & \cellcolor[HTML]{F6F5F1}\textemdash & \cellcolor[HTML]{F9DFC4}\added{212} & 68.2 & 26.5 & 38.2 & 1,156 & \cellcolor[HTML]{F5CAA8}\added{873} & \cellcolor[HTML]{F6F5F1}\textemdash & \cellcolor[HTML]{F9DDC1}\added{283} & 90.1 & 24.5 & 38.5 \\
    \midrule
    \multirow{2}{*}{GPTZero$^{\dagger}$} & R & 1,001 & \cellcolor[HTML]{E6ECDB}\added{629} & \cellcolor[HTML]{F6F5F1}\textemdash & \cellcolor[HTML]{F5CAA8}\added{372} & 63.3 & 62.8 & 63.1 & \multicolumn{7}{c}{\textemdash} \\
     & H & 1,122 & \cellcolor[HTML]{F5CBA9}\added{365} & \cellcolor[HTML]{F6F5F1}\textemdash & \cellcolor[HTML]{DEE9D9}\added{757} & 67.1 & 67.5 & 67.3 & \multicolumn{7}{c}{\textemdash} \\
    \midrule
    \multirow{3}{*}{Ours} & R & 1,015 & \cellcolor[HTML]{BED9C7}\added{\textbf{923}} & \cellcolor[HTML]{C1DBCA}\added{23} & \cellcolor[HTML]{E1EADA}\added{69} & 99.8 & 90.9 & \textbf{95.2} & 1,023 & \cellcolor[HTML]{B9D6C4}\added{\textbf{965}} & \cellcolor[HTML]{C1DBCA}\added{24} & \cellcolor[HTML]{C7DECE}\added{34} & 99.8 & 94.3 & \textbf{97.0} \\
     & P & 245 & \cellcolor[HTML]{B2D1BF}\added{0} & \cellcolor[HTML]{B2D1BF}\added{\textbf{245}} & \cellcolor[HTML]{B2D1BF}\added{0} & 91.4 & 100.0 & \textbf{95.5} & 271 & \cellcolor[HTML]{B2D1BF}\added{0} & \cellcolor[HTML]{B2D1BF}\added{\textbf{271}} & \cellcolor[HTML]{B2D1BF}\added{0} & 91.9 & 100.0 & \textbf{95.8} \\
     & H & 1,132 & \cellcolor[HTML]{B3D2C0}\added{2} & \cellcolor[HTML]{B2D1BF}\added{0} & \cellcolor[HTML]{B2D1BF}\added{\textbf{1,130}} & 94.2 & 99.8 & \textbf{97.0} & 1,156 & \cellcolor[HTML]{B3D2C0}\added{2} & \cellcolor[HTML]{B2D1BF}\added{0} & \cellcolor[HTML]{B2D1BF}\added{\textbf{1,154}} & 97.1 & 99.8 & \textbf{98.5} \\
    \bottomrule
  \end{tabular}}
    \caption{Label-level results on PDF and BibTeX inputs.
    $^\dagger$ denotes binary auditors.}
  \label{tab:label-level}
\end{table*}
\vspace{-2pt}
\begin{table*}[t]
  \centering
  \footnotesize
\setlength{\tabcolsep}{3.0pt}
    \renewcommand{\arraystretch}{1.08}
    \setlength{\aboverulesep}{0.30ex}
    \setlength{\belowrulesep}{0.35ex}
  \begin{tabular}{lll|c|ccc|ccccccc}
    \toprule
    Method & Input & Metric & R & P1 & P3 & P-avg & H1 & H2 & H3 & H4 & H5 & H6 & H-avg \\
    \midrule
   \multirow{4}{*}{GPT-5.5}
      & \multirow{2}{*}{PDF} & TPR & \cellcolor[HTML]{B9D6C4} 94.3 & \cellcolor[HTML]{FBE9D4} 40.2 & \cellcolor[HTML]{F5CBA9} 1.3 & \cellcolor[HTML]{F8DBBE} \added{20.7} & \cellcolor[HTML]{D7E7D7} 72.2 & \cellcolor[HTML]{D0E4D4} 77.2 & \cellcolor[HTML]{BFDAC8} 89.7 & \cellcolor[HTML]{C2DCCB} 87.2 & \cellcolor[HTML]{B7D4C2} 96.4 & \cellcolor[HTML]{B8D5C3} 95.5 & \cellcolor[HTML]{C3DCCB} 86.4 \\
   
      &  & FPR & \cellcolor[HTML]{FBEBD7} 11.5 & \cellcolor[HTML]{B3D1BF} 0.1 & \cellcolor[HTML]{B6D4C2} 0.7 & \cellcolor[HTML]{B5D3C1} 0.4 & \cellcolor[HTML]{BCD8C6} 1.6 & \cellcolor[HTML]{BDD8C7} 1.7 & \cellcolor[HTML]{B5D3C1} 0.5 & \cellcolor[HTML]{B5D3C1} 0.4 & \cellcolor[HTML]{B3D1BF} 0.1 & \cellcolor[HTML]{E2EADA} 6.9 & \cellcolor[HTML]{BED9C8} 1.9 \\
   
      & \multirow{2}{*}{BibTeX} & TPR & \cellcolor[HTML]{BAD6C5} 93.6 & \cellcolor[HTML]{CEE3D3} 78.0 & \cellcolor[HTML]{F6CDAC} 3.9 & \cellcolor[HTML]{FBEAD5} 41.0 & \cellcolor[HTML]{BED9C8} 90.5 & \cellcolor[HTML]{C0DAC9} 88.9 & \cellcolor[HTML]{BCD7C6} 92.4 & \cellcolor[HTML]{B4D2C0} \textbf{98.5} & \cellcolor[HTML]{B4D2C0} 98.5 & \cellcolor[HTML]{BAD6C4} 94.0 & \cellcolor[HTML]{BAD6C4} 93.8 \\
   
      &  & FPR & \cellcolor[HTML]{CBE1D1} 3.9 & \cellcolor[HTML]{B3D1BF} 0.1 & \cellcolor[HTML]{B3D2C0} 0.2 & \cellcolor[HTML]{B3D2C0} 0.2 & \cellcolor[HTML]{BAD6C4} 1.2 & \cellcolor[HTML]{C1DBCA} 2.4 & \cellcolor[HTML]{B3D2C0} 0.2 & \cellcolor[HTML]{B4D2C0} 0.3 & \cellcolor[HTML]{B2D1BF} 0.0 & \cellcolor[HTML]{E8ECDB} 7.6 & \cellcolor[HTML]{BFD9C8} 2.0 \\
    \midrule
   \multirow{4}{*}{Claude 4.7 Opus}
      & \multirow{2}{*}{PDF} & TPR & \cellcolor[HTML]{BBD7C5} \added{93.4} & \cellcolor[HTML]{F9DCC0} \added{24.1} & \cellcolor[HTML]{F7D5B6} 13.3 & \cellcolor[HTML]{F8D8BB} \added{18.7} & \cellcolor[HTML]{F7F0DF} 53.0 & \cellcolor[HTML]{FCF0DE} \added{49.2} & \cellcolor[HTML]{C0DAC9} \added{90.2} & \cellcolor[HTML]{BED9C8} \added{91.8} & \cellcolor[HTML]{B8D5C3} \added{96.9} & \cellcolor[HTML]{CCE2D2} \added{82.1} & \cellcolor[HTML]{D0E4D4} \added{77.2} \\
   
      &  & FPR & \cellcolor[HTML]{F7D3B4} \added{17.2} & \cellcolor[HTML]{B3D1BF} 0.1 & \cellcolor[HTML]{D2E6D6} 5.0 & \cellcolor[HTML]{C2DBCA} 2.5 & \cellcolor[HTML]{BCD8C6} 1.6 & \cellcolor[HTML]{B3D1BF} \added{0.1} & \cellcolor[HTML]{B5D3C1} \added{0.5} & \cellcolor[HTML]{B5D3C1} \added{0.5} & \cellcolor[HTML]{B2D1BF} \added{0.0} & \cellcolor[HTML]{D7E7D7} \added{5.5} & \cellcolor[HTML]{BCD8C6} \added{1.4} \\
   
      & \multirow{2}{*}{BibTeX} & TPR & \cellcolor[HTML]{BED9C7} \added{92.8} & \cellcolor[HTML]{F9F0DF} \added{52.7} & \cellcolor[HTML]{F7D6B8} \added{16.1} & \cellcolor[HTML]{FAE5CC} \added{34.4} & \cellcolor[HTML]{DAE8D8} \added{75.0} & \cellcolor[HTML]{F7F0DF} \added{54.5} & \cellcolor[HTML]{BCD8C6} \added{94.9} & \cellcolor[HTML]{BCD8C6} \added{94.9} & \cellcolor[HTML]{B3D1BF} \added{99.5} & \cellcolor[HTML]{C2DCCB} \added{89.8} & \cellcolor[HTML]{C9E0D0} \added{84.8} \\
   
      &  & FPR & \cellcolor[HTML]{FBEAD4} \added{10.5} & \cellcolor[HTML]{B3D2C0} 0.2 & \cellcolor[HTML]{CFE4D4} \added{4.3} & \cellcolor[HTML]{C1DBCA} \added{2.2} & \cellcolor[HTML]{B8D5C3} \added{0.9} & \cellcolor[HTML]{B5D3C1} \added{0.4} & \cellcolor[HTML]{B4D2C0} \added{0.3} & \cellcolor[HTML]{B5D3C1} \added{0.4} & \cellcolor[HTML]{B3D1BF} \added{0.1} & \cellcolor[HTML]{DFE9D9} \added{6.4} & \cellcolor[HTML]{BCD8C6} \added{1.4} \\
    \midrule
   \multirow{4}{*}{Gemini 3.1 Pro}
      & \multirow{2}{*}{PDF} & TPR & \cellcolor[HTML]{BFD9C8} 90.0 & \cellcolor[HTML]{F8D8BA} 17.2 & \cellcolor[HTML]{F5CCAA} 2.5 & \cellcolor[HTML]{F7D2B2} 9.9 & \cellcolor[HTML]{F9DFC4} 26.3 & \cellcolor[HTML]{F9DCC0} 22.8 & \cellcolor[HTML]{F0EEDD} 56.7 & \cellcolor[HTML]{E4EBDA} 64.1 & \cellcolor[HTML]{E3EBDA} 64.6 & \cellcolor[HTML]{FAE7D1} 37.2 & \cellcolor[HTML]{FBEDDA} 45.3 \\
   
      &  & FPR & \cellcolor[HTML]{F5CAA8} 41.5 & \cellcolor[HTML]{B6D4C2} 0.7 & \cellcolor[HTML]{CFE4D4} 4.5 & \cellcolor[HTML]{C3DCCB} 2.6 & \cellcolor[HTML]{D0E5D5} \added{4.6} & \cellcolor[HTML]{B5D3C1} 0.5 & \cellcolor[HTML]{B6D4C2} 0.7 & \cellcolor[HTML]{B8D5C3} 0.9 & \cellcolor[HTML]{B2D1BF} \added{0.0} & \cellcolor[HTML]{C3DCCB} 2.6 & \cellcolor[HTML]{BCD8C6} 1.6 \\
   
      & \multirow{2}{*}{BibTeX} & TPR & \cellcolor[HTML]{C1DBC9} 88.5 & \cellcolor[HTML]{F8DABC} 19.8 & \cellcolor[HTML]{F6D0AF} 7.2 & \cellcolor[HTML]{F7D5B6} 13.5 & \cellcolor[HTML]{F8D9BC} 19.5 & \cellcolor[HTML]{F9DFC3} 25.8 & \cellcolor[HTML]{F6F0DF} 53.3 & \cellcolor[HTML]{DBE8D8} 69.7 & \cellcolor[HTML]{FCEFDC} 47.0 & \cellcolor[HTML]{F8D9BC} 19.3 & \cellcolor[HTML]{FBE9D3} 39.1 \\
   
      &  & FPR & \cellcolor[HTML]{F5CAA8} 46.3 & \cellcolor[HTML]{B3D2C0} 0.2 & \cellcolor[HTML]{FBE9D4} 12.0 & \cellcolor[HTML]{DBE8D8} 6.1 & \cellcolor[HTML]{C1DBCA} 2.4 & \cellcolor[HTML]{B5D3C1} 0.4 & \cellcolor[HTML]{B7D4C3} 0.8 & \cellcolor[HTML]{B7D4C3} 0.8 & \cellcolor[HTML]{B2D1BF} 0.0 & \cellcolor[HTML]{B8D5C3} 0.9 & \cellcolor[HTML]{B8D5C3} 0.9 \\
    \midrule
   \multirow{2}{*}{GPTZero$^{\dagger}$}
      & \multirow{2}{*}{PDF} & TPR & \cellcolor[HTML]{E8ECDB} 62.8 & --- & --- & --- & \cellcolor[HTML]{FAF1E0} 51.0 & \cellcolor[HTML]{FAE5CD} 34.5 & --- & \cellcolor[HTML]{D5E7D7} 72.8 & \cellcolor[HTML]{FAE4CC} 33.3 & \cellcolor[HTML]{FBE8D1} 37.8 & \cellcolor[HTML]{FCEEDB} 45.9 \\
   
      &  & FPR & \cellcolor[HTML]{F5CAA8} 32.5 & --- & --- & --- & \cellcolor[HTML]{C8DFCF} 3.5 & \cellcolor[HTML]{DBE8D8} 6.1 & --- & \cellcolor[HTML]{F5CAA8} 33.6 & \cellcolor[HTML]{D5E7D7} 5.4 & \cellcolor[HTML]{FBEBD6} 11.6 & \cellcolor[HTML]{FBE9D4} 12.0 \\
    \midrule
   \multirow{4}{*}{Ours}
      & \multirow{2}{*}{PDF} & TPR & \cellcolor[HTML]{BED9C7} \added{90.9} & \cellcolor[HTML]{B2D1BF} \textbf{100.0} & \cellcolor[HTML]{B3D1BF} \textbf{99.4} & \cellcolor[HTML]{B2D1BF} \textbf{99.7} & \cellcolor[HTML]{B2D1BF} \textbf{100.0} & \cellcolor[HTML]{B3D2C0} \textbf{99.0} & \cellcolor[HTML]{B3D1BF} \textbf{99.5} & \cellcolor[HTML]{B8D5C3} \textbf{95.4} & \cellcolor[HTML]{B2D1BF} \textbf{100.0} & \cellcolor[HTML]{B2D1BF} \textbf{100.0} & \cellcolor[HTML]{B3D2C0} \textbf{99.0} \\
   
      &  & FPR & \cellcolor[HTML]{B3D1BF} \textbf{0.1} & \cellcolor[HTML]{B6D3C2} 0.6 & \cellcolor[HTML]{B5D3C1} \textbf{0.4} & \cellcolor[HTML]{B5D3C1} 0.5 & \cellcolor[HTML]{B9D6C4} \textbf{1.1} & \cellcolor[HTML]{B8D5C3} 1.0 & \cellcolor[HTML]{B4D2C0} \textbf{0.3} & \cellcolor[HTML]{B2D1BF} \textbf{0.0} & \cellcolor[HTML]{B9D6C4} 1.1 & \cellcolor[HTML]{B3D1BF} \textbf{0.1} & \cellcolor[HTML]{B6D3C2} \textbf{0.6} \\
   
      & \multirow{2}{*}{BibTeX} & TPR & \cellcolor[HTML]{B9D6C4} \textbf{94.3} & \cellcolor[HTML]{B2D1BF} \textbf{100.0} & \cellcolor[HTML]{B3D1BF} \textbf{99.4} & \cellcolor[HTML]{B2D1BF} \textbf{99.7} & \cellcolor[HTML]{B2D1BF} \textbf{100.0} & \cellcolor[HTML]{B3D2C0} \textbf{99.0} & \cellcolor[HTML]{B3D1BF} \textbf{99.5} & \cellcolor[HTML]{B8D5C3} 95.4 & \cellcolor[HTML]{B2D1BF} \textbf{100.0} & \cellcolor[HTML]{B2D1BF} \textbf{100.0} & \cellcolor[HTML]{B3D2C0} \textbf{99.0} \\
   
      &  & FPR & \cellcolor[HTML]{B3D1BF} \textbf{0.1} & \cellcolor[HTML]{B6D3C2} \textbf{0.6} & \cellcolor[HTML]{B5D3C1} \textbf{0.4} & \cellcolor[HTML]{B5D3C1} \textbf{0.5} & \cellcolor[HTML]{B5D3C1} \textbf{0.5} & \cellcolor[HTML]{B6D4C2} \textbf{0.7} & \cellcolor[HTML]{B4D2C0} \textbf{0.3} & \cellcolor[HTML]{B2D1BF} \textbf{0.0} & \cellcolor[HTML]{B4D2C0} \textbf{0.3} & \cellcolor[HTML]{B3D1BF} \textbf{0.1} & \cellcolor[HTML]{B4D2C0} \textbf{0.3} \\
    \bottomrule
  \end{tabular}
  \caption{Per-subtype TPR and \added{one-vs-rest} FPR (\%).}
  \label{tab:subtype-acc}
\end{table*}

\subsection{Experiment Setup}
\label{sec:exps:setup}

\noindent\textbf{Datasets and Input Modes.}
We evaluate on \added{two primary} corpora introduced in Section~\ref{sec:data}: a synthetic benchmark of $2{,}450$ citations covering the $11$ taxonomy codes except \textsc{P2}, and a $957$-citation real-world set drawn from $647$ ICLR 2026 and $41$ another anonymous conference desk-rejected submissions that venue chairs flagged as fabricated references.
Synthetic-benchmark citations are rendered under five bibliography styles spanning single-column (\texttt{plain}, ICLR) and two-column (IEEE, ACM Reference Format, Springer LNCS) layouts.
Each system is run under two input modes: \emph{PDF input} on the rendered benchmark PDF ($N = 2{,}392$ after excluding $58$ render-omitted citations) and \emph{BibTeX input} on the source \texttt{.bib} entries ($N = 2{,}450$).
\added{We also evaluate transfer beyond computer science on BIOMED-1K (Appendix~\ref{sec:exps:crossdomain}).}

\noindent\textbf{Baselines.}
We compare \sys against frontier AI chatbots and existing citation auditors:
GPT-5.5 Thinking~\cite{openai2026chatgpt}, Claude 4.7 Opus Adaptive Thinking~\cite{anthropic2026claude}, and Gemini 3.1 Pro~\citep{google2026gemini}, prompted with the same audit prompt, as well as Hallucinator~\citep{hallucinator} and GPTZero~\citep{gptzero}.
Neither Hallucinator nor GPTZero exposes a \textsc{Potential} prediction class, so we score them as binary \textsc{Real}-vs-\textsc{Hallucinated} classifiers. GPTZero further accepts only PDF input.

\noindent\textbf{Evaluation Metrics.}
\added{We report both three-class verdict accuracy
and fine-grained diagnostic accuracy.
The former evaluates the final
\textsc{Real}/\textsc{Potential}/\textsc{Hallucinated}
verdict, while the latter evaluates the predicted
diagnostic code after merging \textsc{R1}--\textsc{R3}
into \textsc{R}.} For each of the three verdict classes, we additionally report one-vs-rest precision, recall, and $F_1$. At the subtype level, we report in-bucket TPR (recall) and one-vs-rest FPR.

\subsection{Main Verification Results}
\label{sec:exps:main}

\noindent\textbf{Label-level Performance.}
\added{As shown in~\autoref{tab:label-level}, \sys consistently
outperforms all three-way baselines across all citation
categories on both BibTeX and PDF inputs.
The improvement is particularly pronounced for
\textsc{Potential}, where \sys achieves an $F_1$ of
$95.8$ on BibTeX, compared with $43.8$ for the strongest
chatbot, GPT-5.5.
This substantial margin highlights the effectiveness of
our fine-grained verification framework in handling
ambiguous citations that require more than straightforward
metadata matching.
Moreover, \sys maintains strong performance on PDF inputs,
demonstrating its robustness to the additional challenges
of reference extraction.
Overall, \sys achieves $97.55\%$ three-class accuracy on
BibTeX and $96.07\%$ on PDF.}

\noindent\textbf{Per-subtype Performance.}
\added{As shown in~\autoref{tab:subtype-acc}, \sys
consistently achieves high detection rates across
fine-grained citation categories, with particularly
substantial improvements on ambiguous references.
Notably, \sys achieves a $99.4\%$ TPR on \textsc{P3},
compared with only $3.9\%$ for GPT-5.5, demonstrating
its ability to recognize insufficient bibliographic
evidence rather than incorrectly treating such cases
as definitive errors.
Beyond ambiguous citations, \sys achieves an average
TPR of $99.0\%$ across hallucination subtypes,
compared with $93.8\%$ for GPT-5.5, indicating
strong performance across diverse bibliographic errors.}

\begin{figure}[t]
  \centering
  \includegraphics[width=\columnwidth]{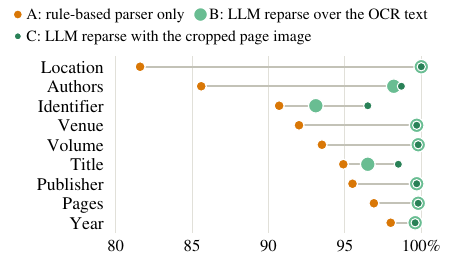}
  \caption{Per-field extraction accuracy (\%) of the three Reference Extractor variants.}
  \label{fig:field-extraction}
\end{figure}

\subsection{Ablations}
\label{sec:exps:ablation}
\noindent\textbf{PDF Extraction.}
We compare three Reference Extractor variants that share the same OCR and reference-segmentation pass and differ only in the parsing step: \textbf{A} rule-based parser only, \textbf{B} adds an LLM reparse over the OCR text, and \textbf{C} attaches the per-entry cropped page image to the same reparse.
As shown in~\autoref{fig:field-extraction}, the LLM reparse step (A to B) is the larger gain, lifting \textsc{Authors} from $85.6$ to $98.2$, \textsc{Location} from $81.6$ to $100.0$, \textsc{Venue} from $92.0$ to $99.7$, and \textsc{Volume} from $93.5$ to $99.8$.
Adding the page image (B to C) is cleaner: \textsc{Title} from $96.5$ to $98.5$ and \textsc{Identifier} from $93.1$ to $96.5$, with the largest gain on the densest layouts.

\begin{figure*}[t]
  \centering
  \includegraphics[width=\textwidth]{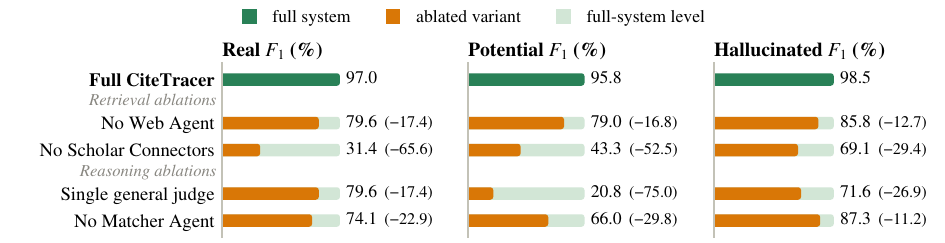}
  \caption{Class $F_1$ (\%) of ablated variants on synthetic BibTeX citations. Retrieval ablations remove the Web Agent or the Scholar Connectors from the cascading evidence collector. \added{Reasoning ablations keep retrieval fixed and either replace the three class-specialist judges with a single general judge or remove the Matcher Agent.}}
  \label{fig:ablation}
\end{figure*}

\vspace{3pt}
\noindent\textbf{Retrieval and Reasoning Ablations.}
Each ablated variant in \autoref{fig:ablation}
removes or replaces one component.
\added{Removing either the Web Agent or Scholar
Connectors substantially degrades performance,
highlighting the complementary roles of web
retrieval and structured scholarly metadata.
More importantly, even with retrieval unchanged,
replacing the class-specialist judges with a single
general judge reduces \textsc{Potential} $F_1$
from $95.8$ to $20.8$.
Removing the Matcher Agent also substantially
weakens \textsc{Real} classification.
These results demonstrate that \sys's effectiveness
depends not only on comprehensive evidence retrieval
but also on specialized reasoning and field-level matching.}

\vspace{-2pt}

\subsection{Real-World Evaluation}

\label{sec:exps:realworld}

We evaluate on two real-world hallucination sets where venue chairs themselves flagged the fabrications. On $807$ citations from $647$ ICLR 2026 desk-rejected submissions, \sys flags $796$ as \textsc{Hallucinated} reference for $\mathbf{98.6\%}$ recall, with the $11$ remaining citations landing in \textsc{Potential}. On $150$ chair-confirmed hallucinated citations from $41$ anonymous conference papers, \sys labels $\mathbf{133}$ as \textsc{Hallucinated} and the remaining $17$ as \textsc{Potential} (author-variant ambiguity), surfacing every confirmed hallucination across both venues. On average each correctly-detected citation triggers $2.24$ distinct error codes, consistent with LLM-fabricated references inventing multiple fields at once.

\section{Conclusion}
\label{sec:conclusion}

\added{We introduce \sys, a citation verification framework
that combines a 12-code taxonomy with cascading evidence
retrieval, field-level matching, and class-specialist judgment.
By distinguishing insufficient evidence from verifiable
bibliographic errors, \sys provides actionable field-level
verdicts beyond binary citation verification.
On the synthetic benchmark, \sys achieves $97.55\%$
three-class accuracy and $97.10\%$ fine-grained diagnostic
accuracy.
It also identifies $97.1\%$ of known fabricated citations
in real-world submissions and achieves $97.50\%$ accuracy
on BIOMED-1K.
Component ablations further demonstrate the contributions
of specialized judgment and field-level matching,
highlighting the effectiveness and generalizability
of our approach.}

% ARR requires a Limitations section after the conclusion and before the
% references; it does not count towards the page limit.
\section*{Limitations}
\added{Our evaluations on computer-science and biomedical
citations demonstrate the cross-domain applicability
of \sys without domain-specific modifications.
The primary limitation lies in retrieval cost and
scalability. Although the cascading design avoids
unnecessary queries, parallel requests to eight Scholar
Connectors can incur substantial API overhead and
trigger rate limits under high concurrency, potentially
reducing evidence coverage.
Future work will explore a coverage-aware connector
router that selects relevant sources based on citation
metadata and API coverage, reducing redundant queries
while maintaining verification accuracy.}

\section*{Ethical Considerations}

\sys is intended to assist researchers and conference
organizers in verifying bibliographic references,
rather than replacing human judgment.
For authors, it provides field-level diagnostics
to help identify and correct citation errors
before submission.
For conference organizers, it supports the
screening of potentially fabricated references
that warrant further investigation.

However, incomplete bibliographic evidence and
parsing errors may lead to incorrect
\textsc{Hallucinated} verdicts, potentially
harming authors' reputations.
Therefore, flagged citations should be manually
verified before any editorial decision,
and bibliographic errors should not be interpreted
as evidence of intentional misconduct.
The use and release of citation data from
conference submissions should also respect
applicable privacy and access restrictions.

% Optional ethics section; it does not count towards the page limit either.

% acl.sty sets \bibliographystyle{acl_natbib}.
\bibliography{main}

\appendix

\newpage
\section{Benchmark Details}
\label{sec:appendix:benchmark}

This appendix expands Section~\ref{sec:data} with the per-code prose, mutation operator schemas, and quality-control protocols that the main paper compresses for space.

\subsection{Per-code Definitions}
\label{sec:appendix:benchmark:codes}

The taxonomy of \autoref{tab:taxonomy} groups $12$ codes into three auditor-facing classes.
The class-level summary in the main paper compresses what each code names. Here we restate the codes in full so an auditor can map a verdict to a concrete auditor action.

\textbf{\textsc{Real}} citations resolve to the intended publication on every field an auditor would normally check.
\textsc{R1} matches the seed BibTeX entry character-for-character.
\textsc{R2} differs only by a normalizable surface variant such as a venue abbreviation, punctuation difference, capitalization change, or initialed author name.
\textsc{R3} replaces a long author list with \emph{et al.} while preserving the correctness of the named authors and the underlying publication.

\textbf{\textsc{Hallucinated}} citations contain field-level bibliographic errors that can be verified against external sources, and each code targets exactly one field so the label identifies the exact correction or auditor action required.
\textsc{H1} corrupts the title through word substitution, paraphrase, or full fabrication.
\textsc{H2} corrupts the author list through addition, deletion, reordering, substitution, or fabrication.
\textsc{H3} preserves title and authors but assigns the work to a venue in which it did not appear.
\textsc{H4} changes the publication year.
\textsc{H5} replaces an identifier with one that either resolves to a different work or fails to resolve.
\textsc{H6} corrupts peripheral metadata (pages, volume, publisher, location) when that metadata can still be checked against an indexed source.

\textbf{\textsc{Potential}} citations cannot be safely resolved by automatic verification alone and should be routed for manual inspection. These cases are not necessarily erroneous, but lack either a stable matching rule or sufficient external evidence for a confident automatic verdict.
\textsc{P1} covers author-name variants, where a citation uses a known nickname, spelling variant, or transliteration variant such as ``Kate'' for ``Katherine'' or ``Mike'' for ``Michael''. Bibliographic records often do not explicitly validate such equivalences, and strict string matching may falsely flag them.
\textsc{P2} marks non-academic sources, including blog posts, GitHub repositories, model release notes, and forum threads, whose citation formats are too diverse to support a uniform bibliographic-index-based judgment.
\textsc{P3} covers peripheral metadata when the relevant field is absent from available bibliographic sources. Because these fields are often less consistently indexed, their absence may reflect incomplete source coverage rather than fabrication.
\added{In particular, \textsc{P3} means the cited core fields identify a real work but the consulted evidence neither supports nor contradicts the supplied peripheral field, whereas \textsc{H6} requires an explicit conflicting record for that field. The distinction concerns available evidence, not whether a generated peripheral value is known to be objectively correct.}

\subsection{Source Selection}
\label{sec:appendix:benchmark:sources}
\begin{table*}[t]
  \centering
  \small
  \begin{tabular}{llp{0.50\linewidth}r}
    \toprule
    Code & Field touched & Operator & Count \\
    \midrule
    R1 & --- & exact seed (no mutation) & 338 \\
    R2 & format only & venue acronym, author initialization, title casing & 342 \\
    R3 & authors & truncate to first author + \emph{et al.} & 343 \\
    \midrule
    P1 & one author & nickname or transliteration variant for one author & 91 \\
    P3 & peripheral & create \texttt{volume}, \texttt{pages}, \texttt{publisher}, or \texttt{location} that no connector indexes & 180 \\
    \midrule
    H1 & title & word substitution, paraphrase, or topic-conditioned fabrication & 200 \\
    H2 & authors & inject a non-existent co-author, drop one, or fully fabricate & 198 \\
    H3 & venue & swap to a plausible but wrong conference or journal & 197 \\
    H4 & year & shift by $\pm$1 to $\pm$5 to a plausible but wrong year & 195 \\
    H5 & DOI / arXiv id & fabricate a syntactically valid identifier that does not resolve, or one that resolves to a different paper & 200 \\
    H6 & peripheral & fabricate a peripheral field that at least one connector \emph{does} index, so the wrongness is verifiable & 166 \\
    \bottomrule
  \end{tabular}
  \caption{Mutation operators and per-code counts in the synthetic benchmark.}
  \label{tab:mutations}
\end{table*}

We extract official BibTeX entries from publicly available open-access bibliographic repositories such as Crossref, DBLP, and arXiv, spanning a broad spectrum of research areas and publication venues.
To control seed quality, we prioritize entries that populate the largest number of bibliographic fields (title, authors, venue, year, identifiers, peripheral metadata), so each seed offers a rich substrate for downstream mutation.
We then apply the per-code mutation operators of \autoref{tab:mutations} to generate the synthetic entries.

\begin{figure*}[t]
  \centering
  \includegraphics[width=\textwidth]{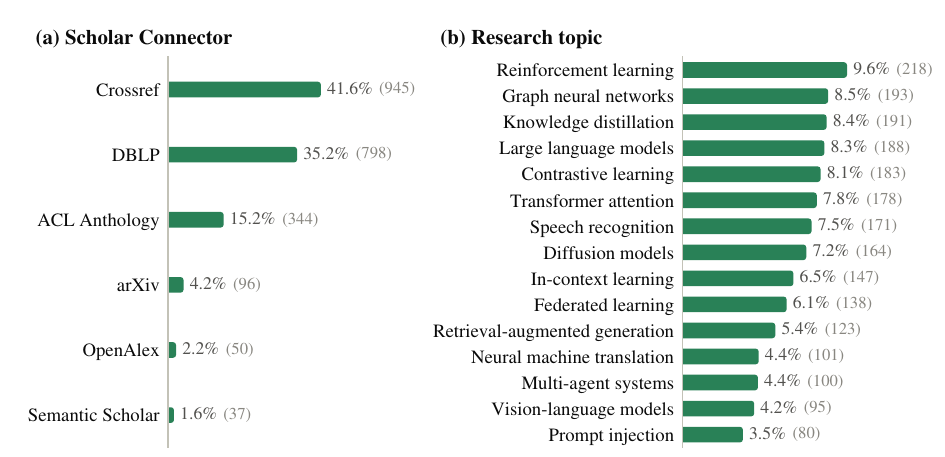}
  \caption{\added{Seed-pool composition of the $2{,}270$ entries included in the plotted pool. Left: distribution over the six Scholar Connectors that returned the canonical record. Right: distribution over the $15$ research topics used to query the connectors.}}
  \label{fig:seed-distribution}
\end{figure*}

\added{\autoref{fig:seed-distribution} summarizes the seed-pool composition for the $2{,}270$ entries shown in the seed-pool plot. The other $180$ \textsc{P3} entries are also synthesized from authentic citations, but are excluded from these plotted source/topic proportions.}
The left panel breaks down seeds by the Scholar Connector that returned the canonical record: Crossref ($41.6\%$) and DBLP ($35.2\%$) together cover three quarters of the pool, ACL Anthology adds $15.2\%$, and the remaining $8.0\%$ is distributed across arXiv, OpenAlex, and Semantic Scholar.
The right panel breaks down seeds by research topic: the $15$ topics span reinforcement learning, graph neural networks, knowledge distillation, large language models, and other major subareas of contemporary AI and machine learning, with no single topic exceeding $9.6\%$ and the smallest topic still contributing $3.5\%$, so no single subarea dominates the benchmark.

\subsection{Per-code Mutation Operators}
\label{sec:appendix:benchmark:mutations}

For every non-\textsc{R1} code we apply a small fixed set of mutation operators that produce exactly the failure mode the code names. Every operator changes a documented set of fields and leaves the rest identical to the seed.
An LLM-driven generator proposes a candidate value for each operator, and a deterministic post-processing step enforces the field boundaries documented in the operator schema.
The \textsc{Potential} class admits operators that no purely surface-text method can recognize. \textsc{P1} substitutes a single author name with a known nickname or transliteration variant, so the citation remains semantically correct yet trips strict matchers. \textsc{P3} fabricates a peripheral field that no public bibliographic source indexes for the cited paper, so the verdict requires recognizing coordinated absence across sources rather than a contradicting source.
\added{The \textsc{P3} operator preserves the real seed's core title, author, venue, and year while adding a plausible peripheral value that cannot be resolved with the collected evidence. It does not fabricate an entirely nonexistent publication.}

Each \textsc{Hallucinated} code targets exactly one bibliographic field, so a wrong title (\textsc{H1}) and a wrong DOI (\textsc{H5}) on otherwise identical seeds produce two distinct benchmark entries and two distinct error modes.

\subsection{Quality Control}
\label{sec:appendix:benchmark:qc}

Every synthetic entry passes three independent checks before it enters the benchmark.
The \emph{round-trip audit} re-runs each operator against its seed and verifies that the resulting diff matches the operator's documented changed fields, and entries that fail the audit are regenerated.
The \emph{verifiability check} confirms that every \textsc{R1} seed resolves on at least one public bibliographic source and that \added{the core fields of each \textsc{P3} entry match a real publication while its synthesized peripheral field cannot be confirmed or contradicted by the consulted sources}, so the \textsc{P3} ground-truth label does not depend on any single source's coverage.
The \emph{author-curated boundary review} hand-inspects every \textsc{P1} citation and confirms that the substituted nickname or transliteration is a recognized variant for the named author rather than a plausible-but-fictional alternative. This protects \textsc{P1} from absorbing \textsc{H2} mutations.
After applying these filters we retain $2{,}450$ taxonomy-labeled instances out of $3{,}100$ collected and synthesized entries.

\section{Efficiency Analysis}
\label{sec:appendix:efficiency}

Across the $2{,}450$-citation BibTeX benchmark, the Cascading Evidence Collector closes $3.6\%$ within seconds via cache hits and non-academic short-circuits, the Field Matcher closes another $61.7\%$ at deterministic rule-based latency with no LLM call, and the remaining $34.7\%$ reach the Class-Specialist Judgers, where the Potential and Hallucinated judges run sequential LLM passes plus external-API cross-checks that account for most of the per-citation latency.
\sys sustains roughly $0.50$ citations per second end-to-end, and the long tail comes primarily from external-API round-trip rather than LLM inference itself.

\section{Implementation Details}
\label{sec:appendix:implementation}

The OCR model $\mathcal{M}_{\mathrm{ocr}}$ uses DeepSeek-OCR 2~\citep{wei2026deepseek} for layout-aware bibliography-region detection and citation-block transcription, and the Parser Agent $\mathcal{A}_{\mathrm{Parser}}$ runs on Kimi K2.5~\citep{kimiteam2026kimik25visualagentic} for the cropped-block reparse and boundary merging.
The Matcher Agent $\mathcal{A}_{\mathrm{Matcher}}$ runs on Qwen3-VL-235B~\citep{bai2025qwen3vltechnicalreport}.
Every LLM call samples at temperature $0$ with a $4{,}096$-token generation cap, and the Cascading Evidence Collector keeps the top-$5$ candidates per connector for downstream adjudication.
The Scholar Connectors $\mathcal{A}_{\mathrm{Scholar}}$ connect to eight academic data sources (arXiv, DBLP, Crossref, Semantic Scholar, OpenAlex, ACL Anthology, Europe~PMC, and PubMed). The URL Fetch step covers direct DOI and arXiv links, and the Web Agent $\mathcal{A}_{\mathrm{Web}}$ uses a general web-search engine for the residual long tail.
By default the pipeline runs three nested layers of parallelism: up to $16$ papers are processed concurrently, within each paper up to $16$ citations are verified in parallel, and within each citation up to $10$ Scholar Connector queries are issued in parallel.

\subsection{Agent Prompts}
\label{sec:appendix:implementation:prompts}

We list the LLM prompts behind the three agents discussed in Section~\ref{sec:method}: the Parser Agent (Reference Extractor), the Matcher Agent (Field Matcher), and the Potential Judger in Class-Specialist Judgers.

\subsubsection{Parser Agent}
\label{sec:appendix:prompts:parser}

The Parser Agent (Section~\ref{sec:method:parser}) takes the OCR transcription of a reference block with the cropped page image and emits a structured citation record.
The system and user prompts the agent uses for the text-only reparse path are:
\vspace{3pt}
\begin{tcolorbox}[title={Parser Agent Prompt}, fontupper=\small, fonttitle=\small,
  boxrule=0.4pt,
  arc=1mm,
  left=1mm,
  right=1mm,
  top=0.6mm,
  bottom=0.6mm,
  boxsep=0.5mm,
  before skip=0mm,
  after skip=0mm,
  halign upper=left,
  valign=top,
  breakable]
\begin{lstlisting}[style=prompt, frame=none]
[System prompt]
You are a strict bibliography parser. Extract structured fields from a raw reference string. Return JSON only.

[User prompt]
Reference:
{raw_text}

Return only valid JSON with exact keys:
{"title": "", "authors": [], "venue": "", "year": null, "volume": "", "pages": "", "publisher": "", "location": "", "doi": "", "arxiv_id": "", "url": ""}

Rules:
- authors must be an array of strings.
- IMPORTANT: If the reference uses 'et al.', 'et al', 'others', or 'and others' to indicate that the author list was truncated, you MUST include that marker as the LAST item in the authors array verbatim (e.g., ['Tom Brown', 'Benjamin Mann', ..., 'et al.']). Do NOT silently drop it -- downstream code uses this marker to detect that the author list is intentionally partial.
- year must be integer or null.
- venue is ONLY the journal or conference name (e.g., 'NeurIPS', 'Nature', 'Proceedings of EMNLP 2021'). Do NOT include pages, volume, issue, location, or publisher in venue.
- volume is the volume number (e.g., '104', '15'). Empty if not present.
- pages is the page range (e.g., '1234-1245'). Recognize 'pp.' as pages. Empty if not present.
- publisher is the publishing organization (e.g., 'Association for Computational Linguistics', 'Springer'). Empty if not present.
- location is the conference location (e.g., 'Online', 'Seoul, Korea'). Empty if not present.
- doi/arxiv_id/url should be plain strings (empty if unknown).
- if uncertain, leave empty string / [] / null.
- do not output explanations.
\end{lstlisting}
\end{tcolorbox}

\subsubsection{Matcher Agent (Field Matcher)}
\label{sec:appendix:prompts:matcher}

The Matcher Agent (Section~\ref{sec:method:matcher}) is invoked when the deterministic rule matcher cannot fully resolve a citation-candidate pair.
For each (citation, candidate) pair the agent emits a per-field verdict on authors, venue, and publisher. The citation-side and candidate-side values for those fields are spliced into the prompt at runtime.
We reproduce its directive, the category labels for each audited field, and the output schema.

\begin{tcolorbox}[title={Matcher Agent Prompt (Field Classifier)}, fontupper=\small, fonttitle=\small,
  boxrule=0.4pt,
  arc=1mm,
  left=1mm,
  right=1mm,
  top=0.6mm,
  bottom=0.6mm,
  boxsep=0.5mm,
  before skip=0mm,
  after skip=0mm,
  halign upper=left,
  valign=top,
  breakable]
\begin{lstlisting}[style=prompt, frame=none]
You are a citation field equivalence classifier. For a single (citation, candidate) pair, produce authoritative verdicts for three fields at once: authors, venue, publisher.

======================================================
AUTHORS -- 4 categories
======================================================
- exact:       every author pair is byte-identical after case normalization.
- r2_initial:  same people; the only first-name difference is single-letter
               initial expansion (e.g., "G. Hao" <-> "Gao Hao").
- p1_variant:  same people (surnames equal), but first-name form differs by
               nickname, multi-letter truncation, transliteration, or
               middle-name add/drop (e.g., "Mike" <-> "Michael",
               "Chao" <-> "Chaowei", "Dmitry P. Vetrov" <-> "Dmitry Vetrov").
- h2_error:    genuinely different people, reordered authors, or count
               mismatch with no et-al marker.

[detailed name-decomposition rules, surname-equality hard rule, count-
 mismatch dedicated rule, and ~50 worked examples are listed in the full
 prompt. Full text is in the released code at
 packages/core/bedrock_agents.py]

======================================================
VENUE -- 3 categories
======================================================
- exact:    byte-identical after case/punctuation normalization.
- alias:    same venue, expressed differently
            (acronym <-> full name; "Proceedings of ..." <-> acronym;
            "EMNLP 2024" <-> "EMNLP"; preprint synonyms; sub-track of the
            same conference; non-English official translation).
- different: distinct venues (e.g., "ACM" vs "ACL", "ICML" vs "NeurIPS").

======================================================
PUBLISHER -- 3 categories (same structure as venue)
======================================================
- exact, alias, different. Aliases include acronym <-> full name
  ("ACM" <-> "Association for Computing Machinery"; "PMLR" <-> "Proceedings
  of Machine Learning Research"; "Springer" <-> "Springer Nature").

======================================================
OUTPUT (JSON only, no other text)
======================================================
For each field, write the "reason" key FIRST (step-by-step evidence walk),
then the "overall" key, which MUST equal the conclusion of that reason.

Required shape:
{
  "authors":   {"reason": "...", "overall": "exact"|"r2_initial"|"p1_variant"|"h2_error"},
  "venue":     {"reason": "...", "overall": "exact"|"alias"|"different"},
  "publisher": {"reason": "...", "overall": "exact"|"alias"|"different"}
}
\end{lstlisting}
\end{tcolorbox}

\subsubsection{Potential Judger}
\label{sec:appendix:prompts:potential}

The Potential Judger (Section~\ref{sec:method:judges}) is the class-specialist agent invoked when the residual field-status profile is consistent with \textsc{Potential} but the system needs to decide between explainable discrepancies (\textsc{P1}/\textsc{P2}/\textsc{P3}) and unexplained errors that escalate to \textsc{Hallucinated}. We additionally include several worked examples as in-context learning demonstrations, and the full set is available in our released code.
\vspace{2pt}
\begin{tcolorbox}[title={Potential Judger Prompt}, fontupper=\small, fonttitle=\small,
  boxrule=0.4pt,
  arc=1mm,
  left=1mm,
  right=1mm,
  top=0.6mm,
  bottom=0.6mm,
  boxsep=0.5mm,
  before skip=0mm,
  after skip=0mm,
  halign upper=left,
  valign=top,
  breakable]
\begin{lstlisting}[style=prompt, frame=none]
You are a citation discrepancy analyst. The citation was NOT a direct match. Your job: determine if the discrepancies are explainable.

## Taxonomy

### POTENTIAL (discrepancies are explainable)
- P1. Author name variant: Name is a plausible nickname/transliteration/spelling variant of the SAME PERSON. The surname sets must still align one-to-one (same count, same people, same order). Only the first-name form differs.
  Valid P1 examples: "Katherine" vs "Kate", "Mike" vs "Michael", "Wei Zhang" vs "William Zhang", "Nando" vs "Fernando", "Shu Yang" vs "Shuang Yang", "Afouras T" vs "Triantafyllos Afouras".
- P2. Non-academic source unverifiable: Citation references a non-academic source whose existence cannot be fully verified through structured databases.
- P3. Insufficient field evidence: Citation provides optional peripheral fields (volume, pages, publisher, location -- ONLY these four) that cannot be confirmed or denied because no candidate source supplies them. The CORE identity fields (title + authors + year + venue) all match the candidate, so the paper is verifiably real -- but one or more of volume/pages/publisher/location carry no external evidence either way.
  Signals: one or more of {volume,pages,publisher,location}_candidate_missing in the issues list while core fields match.
  Valid P3 examples:
    - Citation has volume="35" for a NeurIPS paper; DBLP/CrossRef/arXiv do not supply volume for NeurIPS -> cannot verify. Other fields match -> P3.
    - Citation lists publisher="PMLR" and location="Vancouver, Canada" for an ICML paper; no connector confirms these specific values -> P3.
  Do NOT use P3 when:
    - Any core field (title / authors / year / venue) has a real mismatch.
    - A secondary field has an explicit contradicting value from a candidate (that is H6, not P3).
    - doi_candidate_missing or arxiv_id_candidate_missing is the only issue -- DOI/arxiv_id fall under H5, not P3.

### HALLUCINATED (unexplained errors -> escalate)
*** AUTHOR COUNT / SET MISMATCH IS ALWAYS HALLUCINATED (do NOT explain as P1) ***
These patterns ALWAYS escalate to HALLUCINATED, never P1:
  - Citation has MORE authors than candidate (added author) -> H2
  - Citation has FEWER authors than candidate AND ref has no "et al." -> H2
  - A surname in citation is absent from candidate (or vice versa) -> H2
  - Authors reordered (same set, different order) -> H2
DBLP disambiguation suffixes ("Ting Chen 0007") are NOT a reason to explain away a missing author. The suffix applies to the existing author; a separate added/removed author is still a real discrepancy -> escalate.

If ANY discrepancy cannot be explained by P1/P2/P3, return HALLUCINATED.

## Examples

[Several worked examples for P1, P2, P3, and HALLUCINATED escalation cases are included here as in-context learning demonstrations. Full text is in the released code at packages/core/bedrock_agents.py.]

## Now evaluate:

Citation: Title='{citation_title}' Authors={citation_authors} Venue='{citation_venue}' Year={citation_year} Location='{citation_location}'
Candidate: Title='{candidate_title}' Authors={candidate_authors} Venue='{candidate_venue}' Year={candidate_year} Location='{candidate_location}'

Issues from ValidAgent: {issues}
ValidAgent reason: {valid_reason}

Secondary evidence:
{evidence_lines}

Return JSON only:
{"label": "POTENTIAL"/"HALLUCINATED", "taxonomy": ["P1"/"P2"/"P3"/"H1".."H6"], "reason": "..."}
\end{lstlisting}
\end{tcolorbox}

\section{BibTeX Confusion Matrix}
\label{sec:appendix:confusion}

\begin{figure}[t]
  \centering
  \includegraphics[width=\columnwidth]{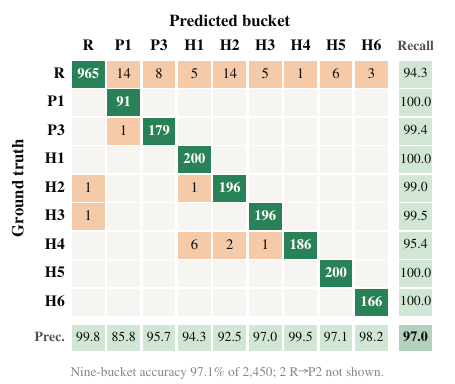}
  \caption{BibTeX confusion matrix of \sys{} \added{(nine displayed buckets). Of $2{,}450$ examples, $2$ true-\textsc{R} citations predicted as \textsc{P2} are omitted because \textsc{P2} has no ground-truth examples in this benchmark. The displayed cells sum to $2{,}448$, while recall and diagnostic accuracy use all $2{,}450$ examples. The margins show precision, recall, and macro-$F_1$.}}
  \label{fig:confusion-bibtex}
\end{figure}

\added{\autoref{fig:confusion-bibtex} shows that $2{,}379$ of the $2{,}450$ BibTeX examples receive the correct nine-bucket code, a fine-grained diagnostic accuracy of $97.10\%$. The \textsc{P1}, \textsc{H1}, \textsc{H5}, and \textsc{H6} rows are perfectly recovered. Among $58$ misclassified \textsc{R} examples, $24$ are predicted as \textsc{P} ($14$ \textsc{P1}, $8$ \textsc{P3}, and $2$ \textsc{P2}) and $34$ as \textsc{H}. The remaining $13$ errors are one \textsc{H2} and one \textsc{H3} citation accepted as \textsc{R}, plus $11$ within-class code confusions: one \textsc{P3}$\rightarrow$\textsc{P1}, one \textsc{H2}$\rightarrow$\textsc{H1}, and nine \textsc{H4} cases assigned another \textsc{H} code.}

\section{\texorpdfstring{\added{Cross-Domain Generalization}}{Cross-Domain Generalization}}
\label{sec:exps:crossdomain}

% Cross-domain results on BIOMED-1K.
\begin{table}[t]
  \centering
  \begingroup
  \small
  \setlength{\tabcolsep}{5pt}
  \begin{tabular}{@{}lccc@{}}
    \toprule
    & \multicolumn{3}{c}{Class $F_1$ (\%)} \\
    \cmidrule(l){2-4}
    Method & Real & Potential & Hallucinated \\
    \midrule
    GPT-5.5 & 93.8 & 92.8 & 94.2 \\
    \sys & \textbf{98.1} & \textbf{94.2} & \textbf{98.7} \\
    \bottomrule
  \end{tabular}
  \caption{\added{Cross-domain evaluation on BIOMED-1K ($1{,}000$ synthetic biomedical citations). We report class-wise $F_1$ scores (\%).}}
  \label{tab:biomed-generalization}
  \endgroup
\end{table}

\noindent\added{We evaluate transfer to biomedical citations on BIOMED-1K, comprising $1{,}000$ citations synthesized using the same taxonomy and mutation operators as the CS benchmark, with real seeds drawn from PubMed, Europe PMC, and bioRxiv. Without changing the verification pipeline or prompts, \sys attains $97.50\%$ three-class verdict accuracy, with a $95\%$ bootstrap percentile interval of $[96.50\%,\,98.40\%]$ from $10{,}000$ resamples. As shown in~\autoref{tab:biomed-generalization}, its $F_1$ exceeds GPT-5.5 on all three verdict classes. These results support transfer to this biomedical benchmark, while leaving other disciplines and naturally occurring citation errors for future evaluation.}

\end{document}